\documentclass[runningheads]{llncs}
\usepackage{makeidx}
\usepackage{graphicx}
\usepackage{amsmath,amssymb} % define this before the line numbering.
\usepackage{color}
\usepackage{hhline}
\usepackage{comment}
\usepackage{subfig}
\usepackage{adjustbox}
\usepackage{multirow}
\usepackage{commath}

\begin{document}
	\pagestyle{headings}
	\mainmatter

	% Replace with your title
	\title{Joint Viewpoint and Keypoint Estimation with Real and Synthetic Data}

	% DO NOT MODIFY these for the draft version that is used for the
	% review process.
	\titlerunning{Joint Viewpoint and Keypoint Estimation with Real and Synthetic Data}
	\authorrunning{Panareda Busto and Gall}
	\author{Pau Panareda Busto\orcidID{0000-0002-2556-2531} \and Juergen Gall\orcidID{0000-0002-9447-3399}}
	\institute{Computer Vision Group, University of Bonn, Germany}

	\maketitle

	\begin{abstract}
		The estimation of viewpoints and keypoints effectively enhance object detection methods by extracting valuable traits of the object instances. While the output of both processes differ, i.e., angles vs.\ list of characteristic points, they indeed share the same focus on how the object is placed in the scene, inducing that there is a certain level of correlation between them. 
		Therefore, we propose a convolutional neural network that jointly computes the viewpoint and keypoints for different object categories. By training both tasks together, each task improves the accuracy of the other.
		Since the labelling of object keypoints is very time consuming for human annotators, we also introduce a new synthetic dataset with automatically generated viewpoint and keypoints annotations. Our proposed network can also be trained on datasets that contain viewpoint and keypoints annotations or only one of them. The experiments show that the proposed approach successfully exploits this implicit correlation between the tasks and outperforms previous techniques that are trained independently.
	\end{abstract}
	
	\section{Introduction}
	\label{sec:introduction}
	
	%Many camera-based applications need to identify and analyse certain object classes for a better undertanding of their surroundings. Most of them, however, rely on the single usage of object detection methods, which become too scarce when we want to extract additional traits from the detected objects. 
	%One task that allows for additional information and has recently attracted a lot of attention is the object's viewpoint estimation with respect to the camera.
	%The current pose of an object not only provides valuable cues for a more robust detection in the coming frames, but also crucial hints about its behaviour.
	%For instance, autonomous vehicles in city scenarios must pay special attention to pedestrian poses in order to reduce the amount of false breaks.
	%Another relevant functionality that supports object detection is the estimation of keypoints. These points of interest are characterisic visual traits of the object class that are shared across all instances and define its appearance.
	%This finer look into the object is thus crucial for estimating its actual orientation, the identification of occluded parts or the recognition of actions.

	Many camera-based applications need to identify and analyse certain object classes for a better understanding of their surroundings. While 2D object detection is often a starting point, it is usually required to extract more detailed information from the detected objects. For instance, 2D keypoints provide additional details regarding the shape of an object and the 3D viewpoint provides the information about the orientation of an object. Both tasks, however, are correlated since the locations of the 2D keypoints depend on the orientation of the object and the 2D keypoints are a cue for the 3D orientation.     	
	%Indeed, both viewpoint and keypoint estimators are partially related and despite of having different outputs, i.e., angles vs. list of keypoints, respectively, their results focus on describing the placement of the object in the scene. That is, instances from a specific class with the same viewpoint tend to have a similar, if not the same, keypoint arragement.
	% \PP{PP: Include figure that show viewpoint and keypoint correlation: for a certain viewpoint, the keypoints have a characteristc layout.}
	%For example, the rear wheels of a front facing vehicle are occluded and exactly one front and one rear wheel are visible if we approach the vehicle from the side.
	In this work, we exploit this implicit correlation and introduce a joint model for 3D viewpoint and 2D keypoint estimation. The proposed network generalises the human pose estimator by Wei et al.~\cite{POSE_Wei16} to multiple objects and it is trained jointly for the two tasks. For the 3D viewpoint estimation, we propose a simple yet effective multi-granular viewpoint classification approach.
	 
	The labelling process for training our network requires nonetheless large amounts of accurate labelled data. While human annotations excel in annotating object instances by bounding boxes, they fail to accurately estimate fine 3D viewpoints~\cite{DA_Busto15}. The same applies for annotating keypoints, which require pixel precision and a correct handling of occlusions.
	In order to alleviate the collection of training data, we propose two solutions.
	Firstly, we design our network such that it can be trained with images from different datasets. The datasets can provide annotations for only viewpoints, keypoints or both.
	Secondly, we make use of synthetic data to increase the amount of training samples since computer generated images are a quick way to collect many training samples, as well as precise ground truth.
	Specifically, we introduce a novel synthetic dataset that includes not only viewpoints, but also accurate keypoints. 
	
	We evaluate our method on 12 popular classes of the \emph{ObjectNet3D}~\cite{DATA_Yu16} dataset, which contains both viewpoint and keypoint annotations. We demonstrate that our method outperforms current well established methods for multi-class viewpoint and keypoint estimation.
	
	\section{Related Work}
	
	\subsection{Viewpoint Estimation}
	We divide viewpoint estimation techniques in two categories: regression methods that compute the pose by optimising in a continuous space~\cite{POSE_Fenzi13,POSE_He14,OBJ_Pepik15} and classification-based methods that simplify the span of viewpoints into a limited set of discrete bins. From the latter, focus of our work,
	Liebelt and Schmid~\cite{OBJ_Liebelt10} optimise a multi-view linear SVM and classify local features to select the winning viewpoint based on a voting approach. Busto et al.~\cite{PANAREDABUSTO201875,DA_Busto15} also use linear SVMs in a one-vs-all multi-class approach to refine coarse annotated viewpoints.
	Su et al.~\cite{SYN_Su15} propose a classification-based CNN model with one bin per degree, i.e. 360 bins for the azimuth angle, and a Gaussian function that spreads the optimisation to neighbouring bins. The training phase uses millions of synthetic samples to compensate the fine viewpoint representation. 
	A coarser discretisations was proposed by Tulsiani and Malik~\cite{POSE_Tulsiani15}, which showed better accuracies when trained on real data.
	Massa et al.~\cite{POSE_Massa16} concluded that classification-based approaches obtain better viewpoint accuracies than regression techniques when jointly trained with an object detector in different popular CNN architectures.
	It has also been shown by Ghodrati et al.~\cite{POSE_Ghodrati14} that these methods using global features extracted from the 2D bounding boxes outperform more complex methods trained on 3D data.
	More recently, Divon and Tal~\cite{POSE_Divon18} introduced a triplet loss to increase the dissimilarity of viewpoints that are far apart.
	Viewpoint estimation can also benefit from 3D object detections, as shown by Kehl et al.~\cite{POSE_Kehl17}, who extended a popular real-time object detector with 3D viewpoint predictions.
	
	Close in spirit to our work, other approaches already used the spatial information of keypoints to estimate accurate viewpoints. Torki and Elgammal~\cite{POSE_Torki11} learn a regression function to compute the azimuth angle of vehicles based on pre-computed local features and their spatial arrangements.	
	Pepik et al.~\cite{OBJ_Pepik12} extend the deformable part model~\cite{OBJ_Pedro10} to 3D objects, optimising at the same time the location and the viewpoint of the object for a fixed number of bins.
	Concretely for hand pose estimation, Zimmermann and Brox~\cite{POSE_Zimmermann17} compute the camera parameters by using keypoint confidence maps as input of the network.
	A deep regression technique is presented by Wu et al.~\cite{POSE_Wu16}, where 2D keypoints are used to estimate the camera parameters after concatenating several fully connected layers.
	Lately, Grabner et al.~\cite{POSE_Grabner18} use the Perspective-n-Point algorithm to extract the viewpoint from a detected 3D bounding box.
	% ~\cite{POSE_Zhou18_2} RGB-D for 3D Keypoint estimation with unsupervised DA among views	
	
	\subsection{Keypoint Estimation}
	
	Research in keypoint estimation has mostly been centred on human articulated poses~\cite{POSE_Belagiannis17,POSE_Chu17,POSE_Tompson15,POSE_Toshev14}.
	In this paper, we expand the CNN model for human pose estimation proposed by Wei et al.~\cite{POSE_Wei16}, who optimised confidence maps for each keypoint. This model appends the later portion of the network several times, i.e., the input of the new stage comes from the output of the previous one, creating larger receptive fields. The deeper the stacked network the more it suppresses ambiguities and better captures the spatial layout of the keypoints.
	Newell et al.~\cite{POSE_Newell16} refined this architecture by adding transposed convolutions at the end of each stage for finer
	confidence maps.
	
	Previous to our work, keypoint estimation in rigid objects has already been in focus.
	Long et al.~\cite{POSE_Long14} initially addressed the capabilities of CNNs for keypoint estimation by dividing the last convolutional layer in smaller cells and training each keypoint as an independent class in a multi-class SVM.
	Moving towards a purely neural network approach, Tulsiani and Malik~\cite{POSE_Tulsiani15} concatenate the spatial information in a fully connected layer and only activate through the network those receptive fields that include the corresponding keypoints. The prediction is further refined with independently computed viewpoints.
	The human pose estimation by~\cite{POSE_Newell16} has been modified by~\cite{POSE_Pavlakos17,POSE_Zhou18} to detected 3D keypoints of multiple rigid classes to consequently estimate the translation and rotation of the object by fitting the keypoints into a shape model.

	\subsection{Synthetic Data}

	Synthetic dataset has been used for many years to easily increase the amount of training samples in object detection tasks~\cite{SYN_Peng15,SYN_Pishchulin11}.
	In recent years, new datasets based on computer generated models with accurate 3D pose information have been proposed.
	For instance, ShapeNet~\cite{DATA_Chang15} provides a large dataset of 3D graphics models for hundreds of object classes. Its drawback comes from the low quality of most of their 3D models.
	From another perspetive, other approaches~\cite{POSE_Wang18,DATA_Xiang14} compensate the lack of photo-realism in synthetic images by aligning 3D models with real samples for accurate 3D pose annotations.

	\section{Joint Viewpoint and Keypoint Estimation}
	\label{technical}
	
\begin{figure}[t]
	\centering
	\includegraphics[width=\textwidth]{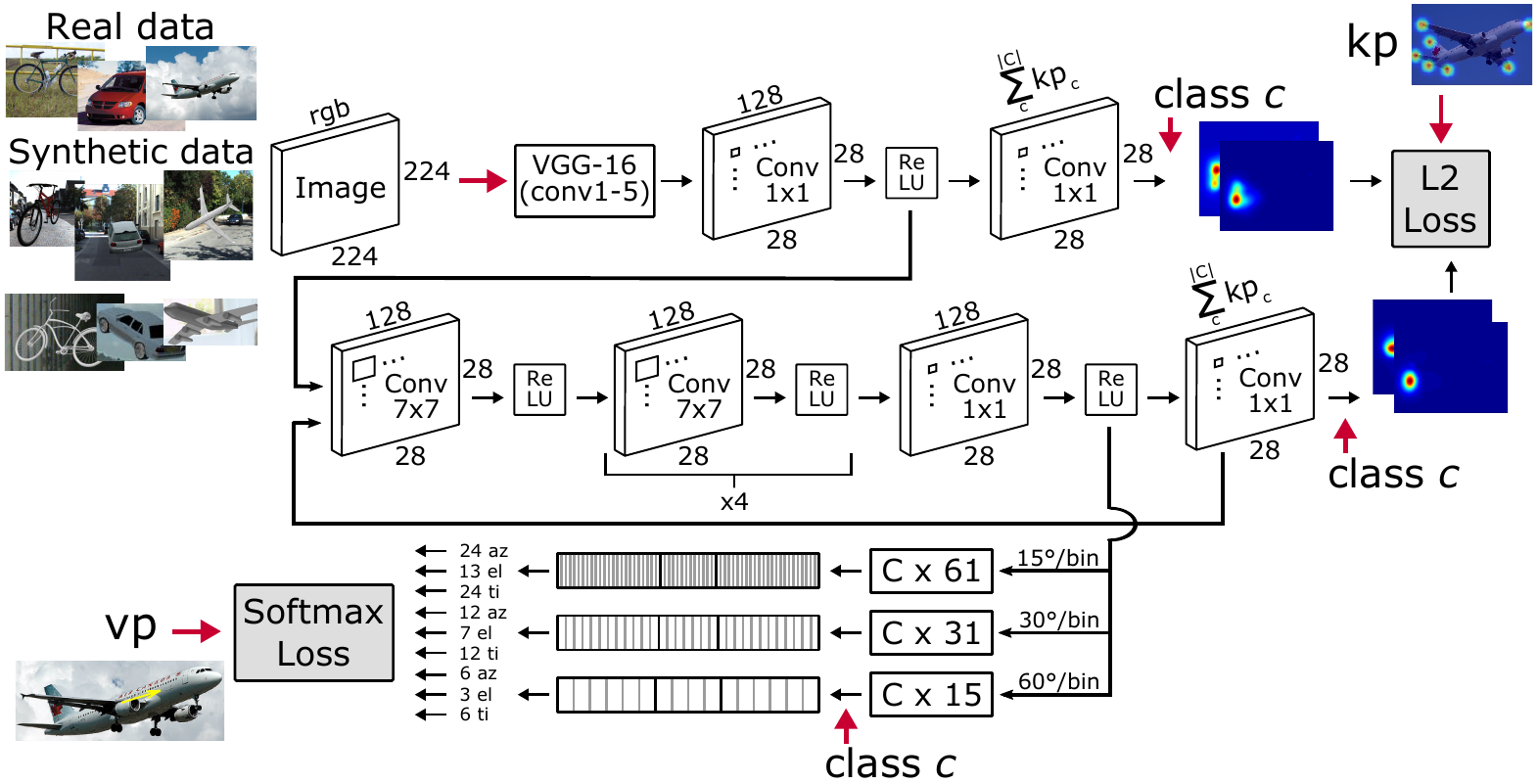}
		\caption{
			Overview of the proposed multi-class CNN for joint viewpoint and keypoint estimation. The network uses a multi-stage architecture. The first row shows the first stage, which predicts for each keypoint per class a heatmap. For the later stages (second row), the features of the first and the previous stage after the last ReLU are used as input. At each stage, an L2-loss is used, which compares the predicted heatmaps for the class of the training sample to the ground truth heatmaps. After the last stage, additional layers for viewpoint estimation are added (third row). We use a multi-resolution loss where fully connected layers map the 128x28x28 features to nine vectors corresponding to three different discretisations (15$^\circ$, 30$^\circ$, 60$^\circ$) of azimuth (az), elevation (el) and tilt (ti).
		}
		\label{fig:cnn}
	\end{figure}

	In this work, we propose a multi-task network that leverages 3D viewpoint and 2D keypoint estimation. We assume that an object has been already detected and our goal is to estimate the keypoints as well as the viewpoint. Our network is trained for all object classes $C = \{c_1, \ldots, c_{|C|}\}$ where the number of keypoints per object class $K_c$ varies.  A second important aspect of the network is that it can be trained on various types of data including real and synthetic data at the same time. Since the data might be annotated for only one of the two tasks, $\mathcal{M}$ denotes the set of training samples with viewpoint and 2D keypoint annotations, $\mathcal{N}$ denotes the set with only viewpoint annotations and $\mathcal{O}$ the set with only keypoint annotations. An overview of the proposed CNN architecture is presented in Figure~\ref{fig:cnn}. We first discuss the parts that are relevant for keypoint estimation.

	%In this section, we describe the overall architecture of our convolutional neural network and the specific details of the viewpoint and keypoint estimation subparts.
	%Given a set of predefined object classes $C = \{c_1, \ldots, c_{|C|}\}$, we aim to learn a multi-class network with training samples $\mathcal{M}$, that contain viewpoint annotations $y_{vp}$ and keypoint annotations $y_{kp}$, $\mathcal{N}$, with only $y_{vp}$, or $\mathcal{O}$, with only $y_{kp}$.
	%An overview of our CNN architecture is presented in Figure~\ref{fig:cnn}.
	
	\subsection{Keypoint Estimation}
	
	The proposed network is a multi-stage architecture with intermediate loss functions after each stage and the first part is similar to the convolutional pose machines~\cite{POSE_Wei16}, which is a multi-stage network for 2D human pose estimation. The cropped image of an detected object is fed to a VGG-16 model~\cite{CNN_Simonyan14} and additional convolutional layers are used to generate heatmaps for each keypoint and each object class. In total, we have $\sum_{c\in C} K_c$ heatmaps, where $K_c$ denotes the number of keypoints of the $c$-th class. Since the object class $c$ is known for an image during training, the $L_2$-loss is computed only for the heatmaps of the corresponding class. At the first stage $s=1$, the loss is therefore given by  
	\begin{equation}\label{eq:keyloss}
		\mathcal{L}_{kp_s} = \sum_{x_i\in\{\mathcal{M},\mathcal{O}\}} \frac{1}{K_{c_i}}\sum_{k=1}^{K_{c_i}}\norm{y_{i,k} - f_s(x_i)_{c,k}}^2_2 \text{,}
	\end{equation}
	where $x_i$ denotes a training sample from the set $\mathcal{M}$ or $\mathcal{O}$ and $f_s(x_i)$ denotes all heatmaps that are predicted for the stage $s$. The estimated heatmap for the $k$-th keypoint of class $c$ is then denoted by $f_s(x_i)_{c,k}$ and $y_{i,k}$ is the corresponding ground-truth heatmap for the training sample $x_i$. The L2-loss is computed over all pixels in the heatmap, but we write $\norm{a-b}^2_2$ instead of $\sum_{\omega\in\Omega}\norm{a(\omega)-b(\omega)}^2_2$. 
	
	As in~\cite{POSE_Wei16}, we do not use one stage but 6 stages. For each stage except of the first one, we use the heatmaps of the previous stage and the feature maps of the first stage after the last ReLU layer as input. Since heatmaps are computed at each stage $s$, we sum the loss functions~\eqref{eq:keyloss} over all stages, i.e., $\sum_s \mathcal{L}_{kp_s}$.

	\subsection{Viewpoint Estimation}
	As shown in Figure~\ref{fig:cnn}, the proposed network not only predicts the 2D keypoints but also the 3D viewpoint encoded by the three angles $\{\phi, \psi, \theta \}$, which denote azimuth ($\phi \in [0^{\circ},360^{\circ}]$), elevation ($\psi \in [-90^{\circ},90^{\circ}]$) and in-plane rotation ($\theta \in [-180^{\circ},180^{\circ}]$), respectively.
	We opt for a classification-based approach to estimate the viewpoints and discretise each angle using a bin size of $15^{\circ}$. We obtain the probabilities for each bin by a fully connected layer and a softmax layer for each angle. The cross-entropy loss for bin size $b=15^{\circ}$ is then given by
	\begin{equation}\label{eq:viewloss}
	\mathcal{L}_{vp_{b}} = \sum_{x_i\in\{\mathcal{M},\mathcal{N}\}} \sum_{v \in \{\phi, \psi, \theta\}} -\log \left ( f_b(x_i)_{c,v,v_i} \right ) \text{,}
	%\left ( \dfrac{\exp(f_{b,v,y_{vp,i}}(x_i))}{\sum^{N_{b,v}}_{n = 1}{\exp(f_{b,v,n}(x_i))}} \right ) \text{,}
	\end{equation}
	where $x_i$ denotes a training sample from the set $\mathcal{M}$ or $\mathcal{N}$, $v_i$ denotes the ground-truth bin for angle $v$ and $f_b(x_i)$ denotes the vector with the bin probabilities for all classes and angles. The estimated probability for the $v_i$-th bin of class $c$ and angle $v$ is then denoted by $f_b(x_i)_{c,v,v_i}$. 
	
	In addition, the network predicts during training the viewpoint for each class for two coarser discretisations of the angles, namely for $60^{\circ}$ and $30^{\circ}$. In this way, the coarse discretisations guide the network to the correct bin of the finer discretisation and improve the accuracy as we will show as part of the experimental evaluation. The multi-task loss for the network is then expressed as
	\begin{equation}
	\mathcal{L} = \sum_s\mathcal{L}_{kp_s} + \sum_b\mathcal{L}_{vp_b}\text{.}  
	\end{equation}
	
\begin{figure}[t]
		\centering
		\includegraphics[width=10cm]{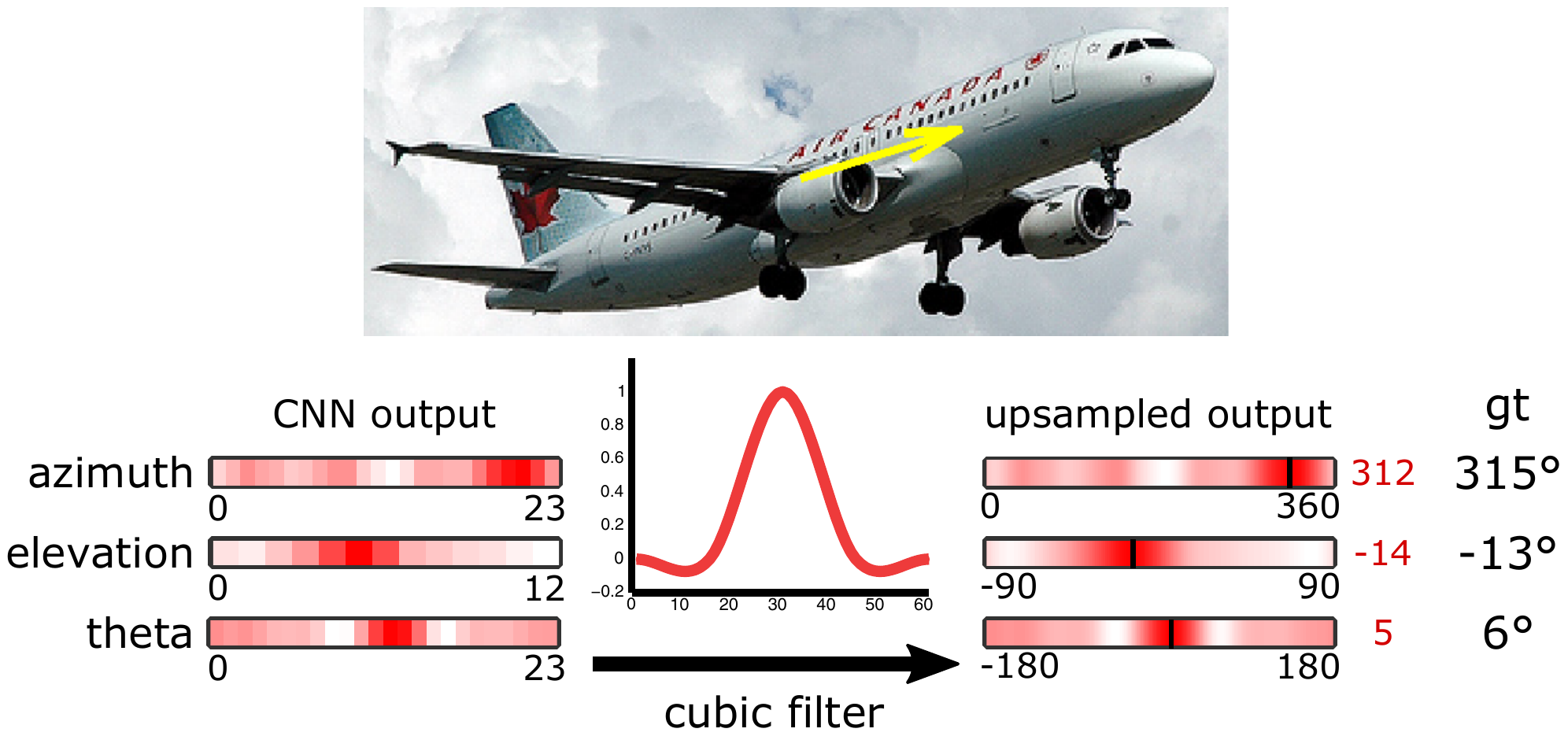}
		\caption{
			Using a cubic filter, the probabilities of the viewpoint quantised at $15^{\circ}$ are upsampled to an angle resolution of $1^{\circ}$. Note that we have 24 bins for azimuth and $\theta$ since they are circular, but only 13 bins for elevation where the 7th bin is centred at zero elevation and the outer bins have only $7.5^{\circ}$.  
		}
		\label{fig:vp}
	\end{figure}

	Since we aim at a finer viewpoint prediction than $15^{\circ}$, we upsample the estimated viewpoint probabilities to an angular resolution of $1^{\circ}$ during inference. To this end, we interpolate the probabilities by applying a cubic filter~\cite{MATH_Keys81} as illustrated in Figure~\ref{fig:vp}. For the azimuth and the in-plane rotation, we convolve the discrete bins as a circular array. %The size of the filter is set to 61 bins.

	\section{Experiments}
	\label{experiments}
	
	In this section we evaluate the performance of our method, denoted as JVK (\emph{Joint Viewpoint and Keypoints}), and compare its results with several popular viewpoint and keypoint estimation algorithms.	We train our network for 12 popular object categories, i.e., $|C|=12$, namely: \emph{airplane}, \emph{bicycle}, \emph{boat}, \emph{bottle}, \emph{bus}, \emph{car}, \emph{chair}, \emph{diningtable}, \emph{motorbike}, \emph{sofa}, \emph{train} and \emph{tvmonitor}. We then evaluate our method on the test images of the \emph{ObjectNet3D}~\cite{DATA_Yu16} dataset. The source code is available at \url{https://github.com/Heliot7/viewpoint-cnn-syn}.
	
	\subsection{Datasets}
	
	\subsubsection{ObjectNet3D~\cite{DATA_Yu16}}
	
	is a large dataset that contains real images of 100 object categories. From all of them, the 12 classes that we selected include not just viewpoints from aligned 3D shapes, but also manually annotated keypoints. 
	The selected subset is evenly separated between training and test data with 11421 and 11327 images, respectively.
	Most of the classes contain between 500 and 1000 samples in every set. The classes \emph{bottle} and \emph{diningtable} are above 1000 samples and \emph{car} above 2000 samples.
	
	\subsubsection{ShapeNet~\cite{DATA_Chang15}}
	
	is a large-scale dataset of 3D shapes whose most relevant subset contains the 12 object categories, providing a considerable amount of models for each class. Although this setting allows for an extensive image dataset with a great variety of object orientations, the low quality of the renderings produce training samples that greatly differ from real images. This dataset only provides 3D viewpoints, automatically generated from the camera parameters in the image rendering.
	For our experiments, we make use of all models and generate 100000 images per class with random camera viewpoints, i.e., 1200000 images in total.
	
	\subsubsection{New Synthetic Data:}
\begin{figure}[p]
		\subfloat[Rendered models with spheres as keypoints]
		{{
			\includegraphics[width=0.6\textwidth]{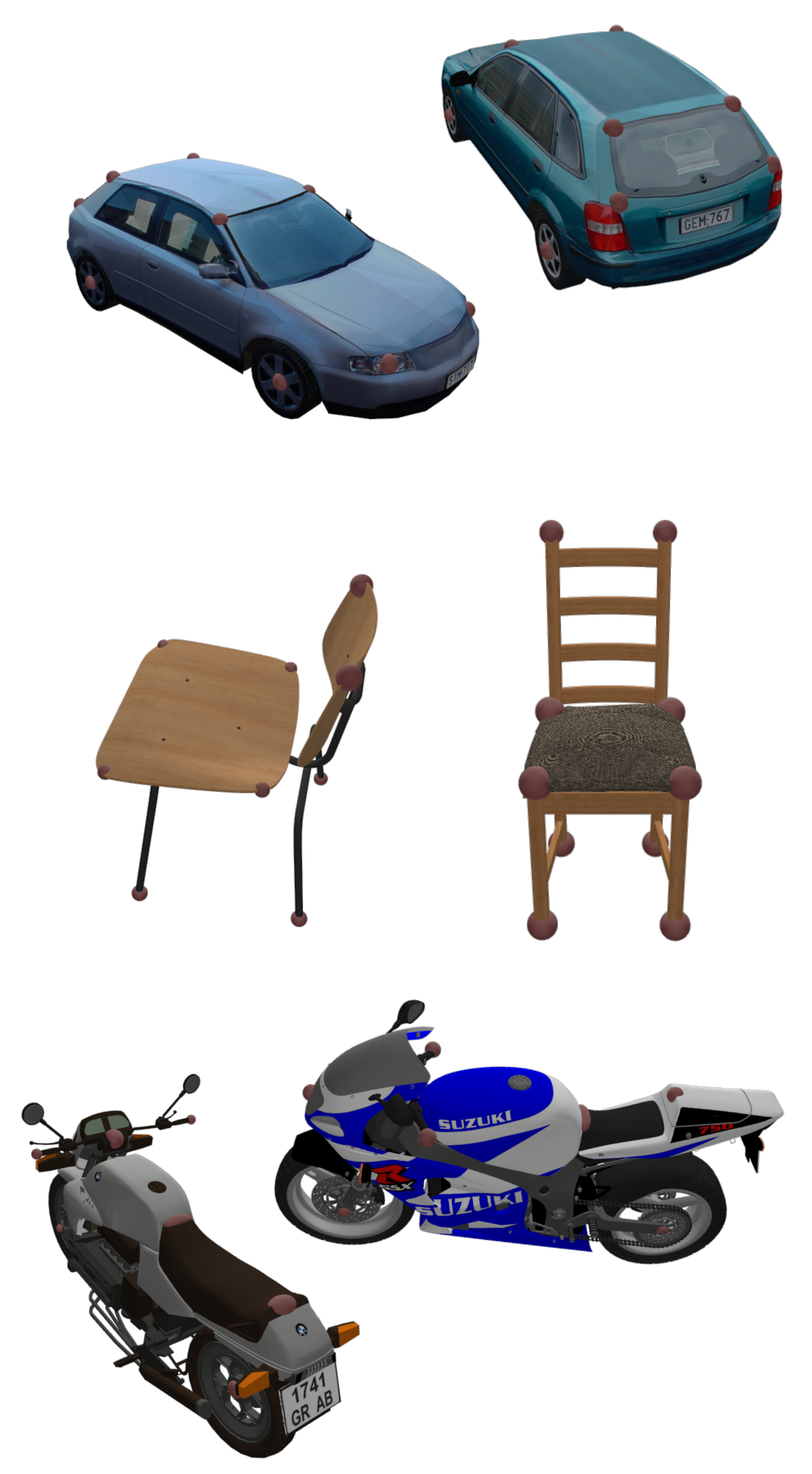}
			\label{fig:syn_a}
		}}
		\subfloat[Generated 2D images]
		{{
			\includegraphics[width=0.37\textwidth]{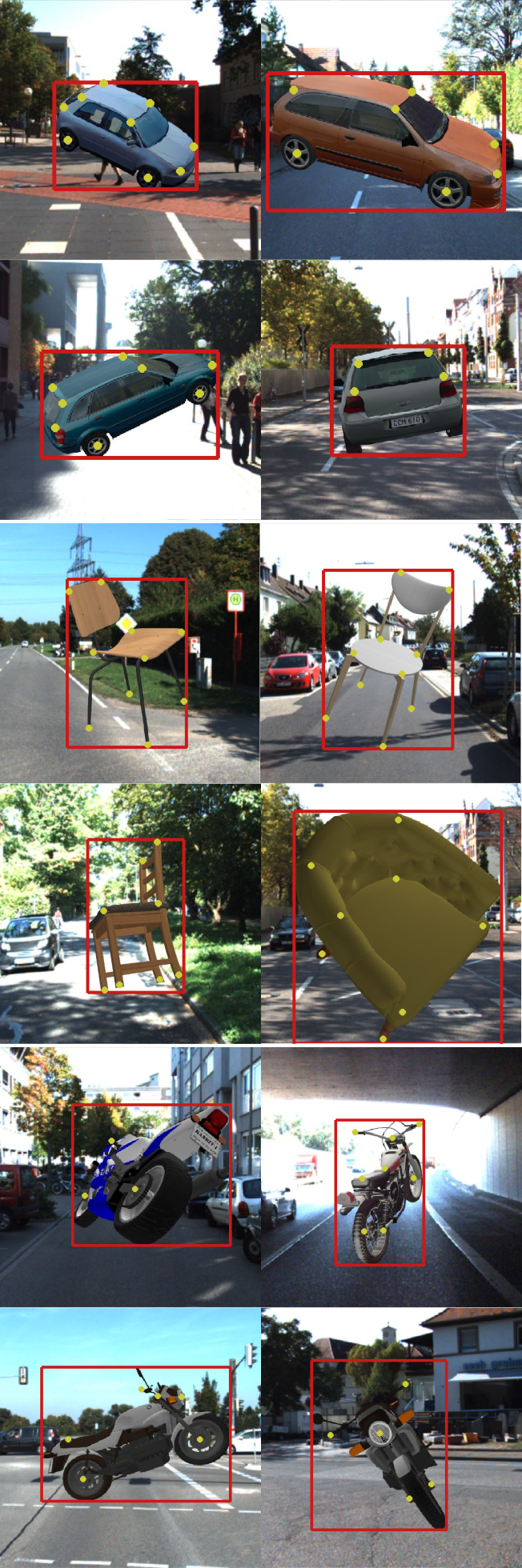}
			\label{fig:syn_b}	
		}}
		\caption{In (a) we show renderings of our graphics models with spheres that represent each keypoint for \emph{cars}, \emph{chairs} and \emph{motorbikes}. In (b) we provide some examples of automatically generated images with their 2D bounding boxes and the projected keypoints that are visible.}
		\label{fig:syn}
	\end{figure}

	In this work, we introduce a new synthetic dataset from 3D graphics models for the 12 object categories.
	For each class, we collect 10 graphics models with higher levels of realism and more detailed meshes compared to ShapeNet.	
	In addition to the 3D viewpoint annotations that are directly extracted from the camera rotation, we go one step further and introduce automatically generated 2D keypoints. In order to easily obtain keypoints from synthetic data, we firstly set deformable spheres in the 3D rendered model locations that we consider to be valid using the keypoints from ObjectNet3D as reference. Figure~\ref{fig:syn_a} shows some 3D graphics models with spheres placed as keypoints.
	Then, we project the centre of each sphere to pixel coordinates for a given camera orientation to create the 2D keypoints. For the projection, we take occlusions into account.  
	%The main difficulty lies in the decision whether the keypoint's visilibity is set to true or false. In order to efficiently solve this problem, we generate two depth maps for both the rendered object and the spheres, separately. We will then consider a keypoint to be visible or not if: (1) at least $p$ depth pixels of the sphere lie closer to the camera than depth values from the rendering in the same xy-coordinates and (2) the distance between both depth values in the same pixel position is below a threshold $d$. In the generation of training data, $p=5$ and $d=0.1\times\text{object length in world coordinates}$. As an example, spheres, that due to the perspective only cover background, will produce occluded keypoints. 
	We generate synthetic data with 10000 samples per class with random orientations. Examples of rendered images are illustrated in Figure~\ref{fig:syn_b} with the 2D bounding boxes and the visible 2D keypoints.
	The resulting images also include a background image from the KITTI dataset~\cite{DATA_Geiger12}.
	%, since we consider most of the object classes to be an active road user. 
	
	\subsection{Network configuration}

	We train the proposed CNN model for a total of 150000 iterations when using only real images for training, 250000 iterations when including one of the two synthetic datasets and 350000 iteration for all 3 datasets. The weight decay is set to 0.0005 and the learning rate to 0.00005, which is multiplied by $0.1$ every 100000 iterations. The input image will be cropped in all experiments to 224x224 pixels while preserving the aspect ratio. The batch contains 20 samples per iteration where we sample uniformly across the datasets if we use more than one for training. In addition, standard data augmentation techniques are employed during the training of the network: flipping, in-plane rotation $[-45^{\circ},45^{\circ}]$, image scaling (0.4,1.0) and translation. However, we only add the transformed image if the intersection over union of the transformed bounding box compared to the original one is above 0.8.
	
	For the test phase, we will extract the samples of each object class using their annotated 2D bounding boxes, i.e., without any prior object detector. We run 5 passes with different scaling factors and average all of them to obtain the final confidence map of keypoints and 3D viewpoints.
	
	From our model, we analyse two modifications. In JVK-KP, we only train the keypoint estimation, ignoring the viewpoint extension. Then, JVK denotes the standard network for both keypoint and viewpoint sections. We also modify the training datasets that we utilise, combining the real samples from ObjectNet3D~\cite{DATA_Yu16} with manually labelled viewpoints and keypoints (Re), ShapeNet~\cite{DATA_Chang15} images with only viewpoints (Sh) and our novel synthetic dataset with generated viewpoints and keypoints (Sy).
	
	\subsection{Keypoint estimation}
	
	To measure the quality of our keypoint localisation, we use the PCK[$\alpha = 0.1$] evaluation introduced by Yang and Ramanan~\cite{POSE_Yang11}.
	An estimated keypoint is valid if the Euclidean distance with respect to the corresponding ground truth is below $\alpha \times max(h,w)$, where $h$ and $w$ are the height and width of the object's bounding box, respectively.
	
	As a baseline, we compare our method with the popular keypoint estimation for rigid objects~\cite{POSE_Tulsiani15} (VpKp). 
	%for different CNN settings, which are also based on the VGG-16 model.
	We report the results of VpKp with 192x192 input resolution (192), 384x384 input resolution (384), both resolutions trained one after the other (192-384) and in a setting where the viewpoint is first estimated for the low resolution and used as input to refine the keypoints for the higher resolution (pLike).

	We report the results in Table~\ref{table:o3d_kps}. Firstly, we observe that JVK-KP (Re), which uses the same real data as in VpKp, already outperforms all variations of VpKp. For instance, our method has $+2.2\%$ accuracy compared to VpKp (pLike). In contrast to VpKp that requires several sequential steps and higher resolutions, we only require a small amount of forward passes of our network with rescaled images. If we compare our modifications, we see a comparable improvement when including synthetic images with only keypoints, JVK-KP (Re-Sy), or only viewpoints, JVK (Re-Sh). This shows the benefits of estimating 3D viewpoint and 2D keypoints jointly. The network trained with all three training datasets (Re-Sy-Sh) obtains the best overall PCK accuracy, which is $+0.7\%$ higher compared to the result without Shapenet (Re-Sy).
	
	\begin{table*}[t]
		\footnotesize
		\begin{center}
			\begin{adjustbox}{max width=\textwidth}
				\begin{tabular}{c|l|cccccccccccc|c}
					% 				\cline{4-14}
					\multicolumn{2}{c|}{ObjectNet3D~\cite{DATA_Yu16} (12 classes)} & aero & bike & boat & bottle & bus & car & chair & dtable & mbike & sofa & train & tv & \multicolumn{1}{c}{Avg.} \\ \hline
					
					\multirow{10}{*}{\begin{tabular}{@{}c@{}}PCK\\ \text{$\alpha=0.1$}\end{tabular}} 
					& VpKp~\cite{POSE_Tulsiani15} (192) & 74.4 & 80.6 & 60.7 & 81.9 & 80.7 & 89.6 & 71.1 & 52.4 & 78.0 & 76.2 & 57.4 & 47.1 & 70.8 \\ \hhline{~~~~~~~~~~~~~~~}
					& VpKp~\cite{POSE_Tulsiani15} (384) & 80.1 & 88.6 & 70.7 & 90.0 & 93.7 & 96.5 & 76.7 & 65.4 & 85.2 & 89.1 & 68.7 & 78.7 & 82.0 \\ \hhline{~~~~~~~~~~~~~~~}
					& VpKp~\cite{POSE_Tulsiani15} (192-384) & 84.1 & 90.0 & 74.4 & 91.3 & 94.4 & 97.5 & 84.9 & 73.3 & 87.4 & 91.0 & 71.3 & 80.1 & 85.0 \\ \hhline{~~~~~~~~~~~~~~~}
					& VpKp~\cite{POSE_Tulsiani15} (pLike) & 82.7 & 90.7 & 69.2 & 92.6 & 95.8 & 95.6 & 89.5 & 76.3 & 85.9 & 92.5 & 72.0 & 80.3 & 85.3 \\ \hhline{~~~~~~~~~~~~~~~}
					& JVK-KP (Re) & 85.7 & 92.7 & 74.8 & \textbf{94.5} & 98.1 & 98.4 & 89.4 & 83.9 & 89.7 & 93.8 & 73.4 & 75.7 & 87.5 \\ \hhline{~~~~~~~~~~~~~~~}
					& JVK (Re-Sh) & 87.9 & 94.7 & 75.3 & 94.3 & \textbf{98.6} & \textbf{98.5} & 89.6 & \textbf{84.5} & 90.6 & \textbf{94.0} & 75.0 & 77.0 & 88.3 \\ \hhline{~~~~~~~~~~~~~~~}
					& JVK-KP (Re-Sy)& 87.7 & 95.2 & 73.6 & 93.9 & 97.8 & \textbf{98.5} & 90.1 & 81.5 & 91.3 & 93.5 & \textbf{75.2} & 83.4 & 88.5 \\ \hhline{~~~~~~~~~~~~~~~}
					& JVK (Re-Sy) & 88.8 & 95.2 & 75.1 & 93.6 & 98.0 & \textbf{98.5} & 90.9 & 83.6 & 91.2 & 93.8 & 73.3 & 82.3 & 88.7 \\ \hhline{~~~~~~~~~~~~~~~}
					& JVK (Re-Sy-Sh) & \textbf{89.5} & \textbf{95.9} & \textbf{77.1} & 93.9 & 98.2 & \textbf{98.5} & \textbf{91.5} & 83.3 & \textbf{93.0} & 93.9 & 74.2 & \textbf{84.0} & \textbf{89.4} \\
					
				\end{tabular}
				
			\end{adjustbox}
		\end{center}
		\caption{Keypoint estimation on the ObjetNet3D dataset~\cite{DATA_Yu16} for 12 object classes. We report the keypoint localisation metric (PCK) introduced by~\cite{POSE_Yang11}.}
		\label{table:o3d_kps}
	\end{table*}
	
	\subsection{Viewpoint estimation}
	
	\begin{table*}[t]
		\footnotesize
		\begin{center}
			\begin{adjustbox}{max width=\textwidth}
				\begin{tabular}{c|l|cccccccccccc|c}
					% 				\cline{4-14}
					\multicolumn{2}{c|}{ObjectNet3D~\cite{DATA_Yu16} (12 classes)} & aero & bike & boat & bottle & bus & car & chair & dtable & mbike & sofa & train & tv & \multicolumn{1}{c}{Avg.} \\ \hline
					\multirow{9}{*}{$Acc{\frac{\pi}{6}}$} & Regression (Re)~\cite{POSE_Massa16} & 79.9 & 81.0 & 66.7 & 93.3 & 92.8 & 96.7 & 90.8 & 79.3 & 83.0 & 96.1 & 94.9 & 89.7 & 87.0 \\ \hhline{~~~~~~~~~~~~~~~}
					& VpKp (Re)~\cite{POSE_Tulsiani15} & 88.7 & 79.4 & 74.3 & 91.7 & 96.7 & 96.3 & 92.2 & 82.3 & 80.8 & 95.4 & 95.7 & 83.1 & 88.0 \\ \hhline{~~~~~~~~~~~~~~~}
					& Render4CNN (Sh)~\cite{SYN_Su15} & 71.0 & 76.1 & 45.1 & 83.7 & 86.3 & 89.9 & 88.5 & 63.0 & 68.4 & 90.4 & 82.3 & 92.3 & 78.1 \\ \hhline{~~~~~~~~~~~~~~~}				
					& Class-15 (Re) & 83.6 & 77.0 & 71.9 & 89.6 & 95.4 & 95.0 & 90.4 & 84.8 & 76.6 & 95.4 & 93.5 & 79.1 & 86.0 \\ \hhline{~~~~~~~~~~~~~~~}
					& Class-15-30-60 (Re) & 85.8 & 81.5 & 71.9 & 92.4 & 96.1 & 95.9 & 92.7 & 85.5 & 81.1 & 95.1 & 94.6 & 83.7 & 87.9 \\ \hhline{~~~~~~~~~~~~~~~}
					& Class-15-30-60 upsamp. (Re) & 86.7 & 82.5 & 73.5 & 92.8 & 95.9 & 96.6 & 93.1 & 85.7 & 81.6 & 96.0 & 94.5 & 85.2 & 88.7 \\ \hhline{~~~~~~~~~~~~~~~}
					& Class (Re-Sy) & 88.4 & 85.8 & 76.5 & 94.5 & 96.9 & 96.8 & 95.6 & 86.5 & 88.5 & 96.5 & 94.3 & 87.5 & 90.7 \\ \hhline{~~~~~~~~~~~~~~~}
					& Class (Re-Sh) & \textbf{91.5} & 85.4 & 80.3 & 94.5 & 97.6 & 97.3 & 97.5 & 86.8 & 86.6 & 97.8 & 95.5 & 89.9 & 91.7 \\ \hhline{~~~~~~~~~~~~~~~}
					& Class (Re-Sy-Sh) & 90.7 & 85.7 & \textbf{81.0} & 93.8 & 98.0 & 97.1 & \textbf{97.9} & 88.3 & 88.5 & 97.9 & 94.6 & 90.3 & 92.0 \\ \hhline{~~~~~~~~~~~~~~~}
					& JVK (Re) & 86.3 & 85.1 & 79.0 & 94.5 & 98.5 & 97.8 & 92.2 & 87.7 & 87.5 & 97.1 & 95.1 & 87.5 & 90.7 \\ \hhline{~~~~~~~~~~~~~~~}
					& JVK (Re-Sy) & 89.8 & \textbf{88.9} & 78.6 & \textbf{95.5} & 98.3 & 97.4 & 93.5 & 87.3 & 90.5 & 97.2 & 94.0 & 88.9 & 91.7 \\ \hhline{~~~~~~~~~~~~~~~}
					& JVK (Re-Sh) & 87.7 & 86.8 & 80.6 & 95.1 & 97.8 & \textbf{98.3} & 96.2 & \textbf{89.2} & \textbf{91.3} & 98.1 & 94.5 & 92.0 & \textbf{92.3} \\ \hhline{~~~~~~~~~~~~~~~}			
					& JVK (Re-Sy-Sh) & 87.8 & 87.0 & 79.8 & 95.0 & \textbf{98.7} & 97.5 & 96.0 & 86.6 & 90.7 & \textbf{98.3} & \textbf{95.8} & \textbf{92.7} & 92.2 \\ \hline
					\multirow{9}{*}{MedError} & Regression (Re) & 13.4 & 16.7 & 18.6 & 8.2 & 4.3 & 4.8 & 9.9 & 11.5 & 16.4 & 9.1 & 6.4 & 13.0 & 11.0 \\ \hhline{~~~~~~~~~~~~~~~}
					& VpKp (Re)~\cite{POSE_Tulsiani15} & 12.2 & 16.0 & 15.4 & 12.7 & 6.8 & 8.9 & 11.6 & 11.1 & 16.8 & 12.3 & 8.0 & 14.0 & 12.2 \\ \hhline{~~~~~~~~~~~~~~~}
					& Render4CNN (Sh)~\cite{SYN_Su15} & 14.9 & 18.6 & 35.5 & 11.4 & 8.2 & 7.5 & 9.5 & 17.4 & 20.1 & 12.9 & 13.0 & 14.6 & 15.3 \\ \hhline{~~~~~~~~~~~~~~~}				
					& Class-15 (Re) & 13.0 & 17.0 & 15.8 & 10.0 & 5.9 & 8.1 & 10.3 & 9.3 & 18.1 & 11.7 & 8.1 & 15.0 & 11.9 \\ \hhline{~~~~~~~~~~~~~~~}
					& Class-15-30-60 (Re) & 11.7 & 15.2 & 15.2 & 9.3 & 5.8 & 8.0 & 9.7 & 9.5 & 17.3 & 11.3 & 8.0 & 14.1 & 11.3 \\ \hhline{~~~~~~~~~~~~~~~}
					& Class-15-30-60 upsamp. (Re) & 9.8 & 13.8 & 13.6 & 8.6 & 4.5 & 5.5 & 7.6 & 7.3 & 15.6 & 9.4 & 6.9 & 13.2 & 9.7 \\ \hhline{~~~~~~~~~~~~~~~}
					& Class (Re-Sy) & 9.0 & 12.5 & 12.5 & 8.0 & 4.2 & 5.1 & 7.2 & 6.8 & 13.0 & 8.6 & 6.1 & 11.4 & 8.7 \\\hhline{~~~~~~~~~~~~~~~}
					& Class (Re-Sh) & \textbf{8.0} & 11.5 & 11.2 & 8.4 & 4.2 & 4.9 & 6.9 & 6.7 & 13.0 & 8.3 & 6.0 & 10.5 & 8.3 \\ \hhline{~~~~~~~~~~~~~~~}
					& Class (Re-Sy-Sh) & 8.3 & 10.9 & \textbf{10.8} & \textbf{7.4} & 4.2 & 4.4 & 6.9 & 6.5 & 12.3 & 7.9 & 6.0 & 10.2 & 8.0 \\ \hhline{~~~~~~~~~~~~~~~}
					& JVK (Re) & 8.5 & 11.2 & 12.3 & 7.5 & 4.1 & \textbf{3.7} & 7.3 & 6.1 & 12.4 & 8.1 & \textbf{5.5} & 9.7 & 8.0 \\ \hhline{~~~~~~~~~~~~~~~}				
					& JVK (Re-Sy) & 8.3 & \textbf{10.0} & 12.0 &\textbf{7.4} & \textbf{3.6} & \textbf{3.7} & \textbf{6.5} & 6.0 & \textbf{11.5} & 7.7 & 5.6 & \textbf{8.9} & \textbf{7.6} \\ \hhline{~~~~~~~~~~~~~~~}
					& JVK (Re-Sh) & 8.4 & 10.4 & 11.2 & \textbf{7.4} & 4.0 & 3.9 & \textbf{6.5} & \textbf{5.6} & 12.1 & \textbf{7.5} & 5.7 & 9.6 & 7.7 \\ \hhline{~~~~~~~~~~~~~~~}
					& JVK (Re-Sy-Sh) & 8.1 & 10.7 & 11.4 & 7.6 & 4.0 & 3.8 & 7.2 & 6.0 & 11.7 & 7.7 & 5.9 & 9.5 & 7.8 \\
					
				\end{tabular}
				
			\end{adjustbox}
		\end{center}
		\caption{Viewpoint estimation on the ObjectNet3D dataset~\cite{DATA_Yu16} from ground truth bounding boxes. We report the percentage of estimated viewpoints with a geodesic error below $\pi/6$ rad ($Acc{\frac{\pi}{6}}$) and the median error (MedError).}
		\label{table:o3d_vp}
	\end{table*}

	We evaluate our viewpoint estimation using two widely used metrics. The first metric~\cite{POSE_Tulsiani15} is the geodesic distance between the ground truth and predicted rotation matrices from $\phi$, $\psi$ and $\theta$, which is given by
	\begin{equation}
	\Delta(R_{gt},R_{pred}) = \frac{||log(R_{gt}^TR_{pred})||_F}{\sqrt{2}}.
	\end{equation}
	The viewpoint is considered to be correct if the distance is below $\frac{\pi}{6}$ rad ($Acc{\frac{\pi}{6}}$). The second measure is the median error (MedError).
	
	For this evaluation against other CNN-based approaches, we take as baseline a standard regression approach by~\cite{POSE_Massa16}, where continous angles are seen as a circular array and represented in $\mathbb{R}^2$. VpKp~\cite{POSE_Tulsiani15} proposes a classification-based viewpoint with also several discretisation levels. Then, Render4CNN~\cite{SYN_Su15} presents a very fine discretisation with Gaussian filters to leverage the neighbouring bins by using millions of synthetic images. Finally, we re-train a VGG-16~\cite{CNN_Simonyan14} model for testing different classification-based configurations (Class): with only one level of discretisation (15$^{\circ}$), our proposed approach with 3 quantisations with $15^{\circ}$, $30^{\circ}$ and $60^{\circ}$, and including the upsampling with cubic filtering (upsamp.).
	
	The evaluation results for all the presented baselines and our configurations are shown in Table~\ref{table:o3d_vp}.
	Generally, we observe that the regression technique obtains similar results compared to other classification-based techniques. However, the cubic interpolation provides a significant reduction in median error and accuracy that favors classification approaches. Compared to the same configuration without upsampling, the error is reduced by $-1.6^{\circ}$ and the accuracy increases by $+0.8\%$. The fine discretisaton of Render4CNN fails to compute robust viewpoints and ends up being the worst performing method by a large margin. Using real images from ObjectNet3D would not solve the problem, since the amount of training samples is too scarce for the large number of bins per angle. Class-15-30-60 outperforms Class-15, showing that learning several angle quantisations at the same time provides better results. When we compare JVK with Class, we observe that including a specific network for keypoint estimation allows for better viewpoint accuracies and reduced angle errors.
	JVK (Re) demonstrates to be superior compared Class upsampling (Re) by $+2\%$ in accuracy and $-1.7^{\circ}$ in the median error. Although the gap is significantly smaller when training the networks with synthetic data, JVK trained with additional synthetic data achieves the best overall results. Specifically, the results of JVK trained on our new synthetic data are comparable to the ones using ShapeNet, but employing 10 times less samples. The better quality and additional labelled data of our dataset play an important role in improving the overall results.
	
	\subsection{Qualitative results}
	
\begin{figure}[p]
		\centering
		\includegraphics[width=\textwidth]{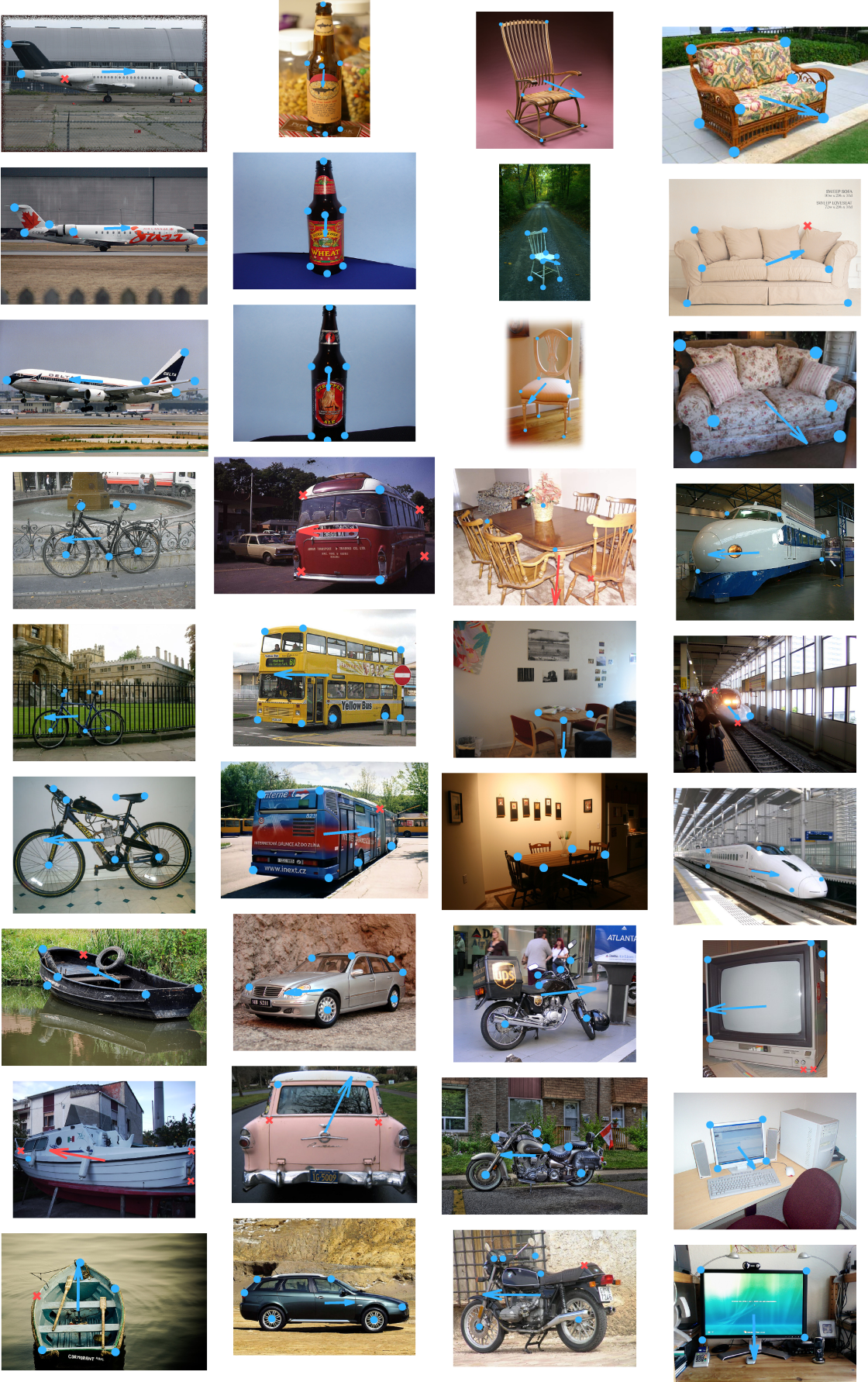}
		\caption{
			Qualitative results for the proposed approach JVK (Re-Sy-Sh). 
			%We draw all detected keypoints that are labelled as visible in their corresponding ground truth. 
			The directional arrow represents the projected 3D viewpoint. Blue (dots) and red (crosses) denote correct and wrong estimations based on the PCK[$\alpha=0.1$] or Acc($\pi/6$) measure, respectively.
		}
		\label{fig:quali}
	\end{figure}

	For completeness, we also show some qualitative results in Figure~\ref{fig:quali}. For each class, we show the results for the first three test images of ObjectNet3D~\cite{DATA_Yu16}. We observe that the predicted 2D keypoints and 3D viewpoints are in alignment. The majority of the few wrongly estimated keypoints and viewpoints are due to lateral symmetries of objects.
	
	\section{Conclusion}

	In this paper we have presented an approach for joint viewpoint and keypoint estimation for multiple rigid object classes. The approach includes a simple yet effective branch for viewpoint estimation with different discretisation levels and cubic upsampling that produce more accurate results. 
	In contrast to previous methods that train a separate approach for each task, we have shown that viewpoint and keypoint estimation benefit from each other. 
	Our approach also handles different kinds of training datasets containing real or synthesized images, as well as datasets where only one of the tasks is annotated. 
	%We have also introduce a new synthetic dataset with automatically generated 2D keypoints that does not require as many training samples as other computer generated datasets, such as ShapeNet.
	We evaluated our approach on ObjectNet3D where it outperforms previous approaches.
	
	\section*{Acknowledgement}
	The work has been supported by the ERC Starting Grant ARCA (677650).

	% \newpage
	\bibliographystyle{splncs04}
	\bibliography{040}

\end{document}